\documentclass{article}
\usepackage[tbtags]{amsmath, nccmath}
\usepackage{microtype}
\usepackage{graphicx}
\usepackage{subfigure}
\usepackage{booktabs} 
\usepackage{setspace}
\usepackage{adjustbox}

\usepackage{hyperref}

\usepackage[accepted]{icml2020}

\icmltitlerunning{Pretrained Generalized Autoregressive  Model with  Adaptive Probabilistic Label Clusters for XMTC}

\begin{document}

\twocolumn[
\icmltitle{Pretrained Generalized Autoregressive Model with Adaptive Probabilistic \\ Label Clusters for Extreme Multi-label Text Classification}



\icmlsetsymbol{equal}{*}

\begin{icmlauthorlist}
\icmlauthor{Hui Ye}{one}
\icmlauthor{Zhiyu Chen}{one}
\icmlauthor{Da-Han Wang}{two}
\icmlauthor{Brian D. Davison}{one}
\end{icmlauthorlist}

\icmlaffiliation{one}{Computer Science and Engineering Department, Lehigh University, PA, USA}
\icmlaffiliation{two}{School of Computer and Information Engineering, Xiamen University of Technology, Fujian, China}

\icmlcorrespondingauthor{Hui Ye }{yehui20080414@gmail.com}

\icmlkeywords{Deep Learning, Text Classification, Extreme labels, Adaptive Probabilistic Label Clusters}

\vskip 0.3in
]



\printAffiliationsAndNotice{} 

\begin{abstract}
Extreme multi-label text classification (XMTC) is a
task for tagging a given text with the most relevant labels from an extremely large label set.
We propose a novel deep learning method called APLC-XLNet. Our approach fine-tunes the recently released generalized autoregressive pretrained model (XLNet) to learn a dense representation for the input text. We propose Adaptive Probabilistic Label Clusters (APLC) to approximate the cross entropy loss by exploiting the unbalanced label distribution to form clusters that explicitly reduce the computational time. Our experiments, carried out on five benchmark datasets, show that our approach has achieved new state-of-the-art
results on four benchmark datasets. Our source code is available publicly at \url{ https://github.com/huiyegit/APLC\_XLNet}.
\end{abstract}

\section{Introduction}
\label{introduction}

Extreme classification is the problem of learning a classifier to annotate each instance with the most relevant labels from an extremely large label set. Extreme classification has found applications in diverse areas, such as
estimation of word representations for millions of words \cite{mikolov2013efficient}, tagging of a Wikipedia article with the most relevant labels \cite{dekel2010multiclass}, and providing a product description or an ad-landing page in Dynamic Search Advertising \cite{jain2019slice}. Extreme multi-label text classification (XMTC) is a fundamental task of extreme classification where both the instances and labels are in text format. 

The first challenge of XMTC is how to get effective features to represent the text. One traditional approach to represent text features is bag-of-words (BOW), where a vector represents the frequency of a word in a predefined vocabulary. Then, the machine learning algorithm is fed with training data that consists of pairs of features and labels to train a classification model. Nevertheless, the traditional methods  based on BOW or its variants, ignoring the location information of the words, cannot capture the contextual and semantic information of the text.

On the other hand, with the recent development of word embedding techniques, deep learning methods have achieved great success for learning text representation from raw text. These effective models include the Convolutional Neural Network (CNN) \cite{kim2014convolutional},  the Recurrent Neural Network (RNN) \cite{liu2016recurrent}, the combination of CNN and RNN \cite{lai2015recurrent}
, the CNN with Attention mechanism \cite{yin2018attentive}, the RNN with Attention mechanism \cite{yang2016hierarchical} and the Transformer \cite{vaswani2017attention, guo2019star}.

Over the last two years, several transfer learning methods and architectures \cite{devlin2019bert,radford2018improving, conneau2019cross, yang2019xlnet} have been proposed, and have achieved state-of-the-art results on a wide range of  natural language processing (NLP) tasks, including question answering \cite{yang2019end}, sentiment analysis \cite{xu2019bert}, text classification \cite{sun2019fine} and information retrieval \cite{chen2020table}. The basic mechanism of transfer learning is to transfer the knowledge of a pretrained model, generally trained on very large corpora, into a new downstream task. The generalized autoregressive pretraining method (XLNet) \cite{yang2019xlnet} represents the latest development in this field.
XLNet adopts the permutation language modeling objective to train the model on several large corpora. Then the pretrained model can be fine-tuned to deal with various NLP tasks. 

Another challenge of XMTC is how to handle the extreme outputs efficiently. In particular, for extreme classification, the distribution of the labels notoriously follows Zipf's Law. Most of the probability mass is covered by only a small fraction of the label set. To train parametric models
for language modeling with very large vocabularies, many methods have been proposed to approximate the softmax efficiently, including Hierarchical Softmax (HSM) \cite{morin2005hierarchical}, Negative Sampling \cite{mikolov2013efficient}, Noise Contrastive Estimation \cite{mnih2012fast, vaswani2013decoding} and Adaptive Softmax \cite{grave2017efficient}. However, these techniques are proposed to deal with multi-class classification, so they can not be applied to XMTC directly. The Probabilistic Label Tree (PLT) \cite{jasinska2016extreme, wydmuch2018no} generalizes the HSM to handle the multi-label classification problem. 

Motivated by these characteristics of the task, we present a novel deep learning approach, Pretrained Generalized Autoregressive  Model with  Adaptive Probabilistic Label Clusters (APLC-XLNet). We fine-tune the  generalized autoregressive pretraining model (XLNet) to learn the powerful text representation to achieve high prediction accuracy. To the best of our knowledge, XLNet is the first time to be applied successfully to the XMTC problem. Inspired by the Adaptive Softmax, we propose the Adaptive Probabilistic Label Clusters (APLC) to approximate the cross entropy loss by exploiting the unbalanced label distribution to form clusters that explicitly reduce computation time. APLC can be flexible enough to achieve the desirable balance between the prediction accuracy and computation time by adjusting its parameters. Furthermore, APLC can be general enough to deal with the extreme classification problem efficiently as the output layer. The experiments, conducted on five datasets, have demonstrated that our approach has achieved new state-of-the-art results on four benchmark datasets. 


\section{Related Work}
\label{section:related_works}
Many effective methods have been proposed for addressing the challenges of XMTC. They can be generally categorized
into two types according to the method used for feature representation. One traditional type is to use the BOW as the feature. It contains three different approaches: one-vs-all approaches, embedding-based approaches and tree-based approaches.
The other type is the modern deep learning approach. Deep learning models have been proposed to learn powerful text representation from the raw text and have shown great success on different NLP tasks.

\textbf{One-vs-all approaches.}
The one-vs-all approaches treat each label independently as a
binary classification problem that learns a classifier for each label. The one-vs-all approaches have been shown to achieve high accuracy, but they suffer
from expensive computational complexity when the number of labels
is very large. 
PDSparse \cite{yen2016pd} learns a separate linear classifier per
label. During training, the classifier
is optimized to distinguish between all the positive labels and a few active negative labels of each training sample. PPDSparse \cite{yen2017ppdsparse} extends PDSparse to be parallelized in large scale distributed settings. 
DiSMEC \cite{babbar2017dismec} presents a distributed and parallel training mechanism, which can  exploit as many computation cores
as are available. In addition, it can reduce the model size by explicitly inducing
sparsity via pruning of spurious weights.
Slice \cite{jain2019slice} trains each label’s classifier over the most confusing negative labels rather than
all the negative labels. This is achieved efficiently by a
novel negative sampling technique.

\textbf{Embedding-based  approaches.} 
Embedding-based approaches project the high-dimensional label space into a low-dimensional one by exploiting label correlations and sparsity.  Embedding methods may pay a heavy price in
terms of prediction accuracy due to the loss of information in the compression phase.
SLEEC \cite{bhatia2015sparse} projects the high-dimensional label space into the low-dimensional
one by preserving the pairwise distances between only the closest label vectors. The regressors can be learned over the low-dimensional label space. AnnexML \cite{tagami2017annexml} partitions data
points by using an approximate k-nearest neighbor graph as weak supervision. Then it employs a pairwise learning-to-rank approach
to learn the low-dimensional label space. 

\textbf{Tree-based  approaches.}
Tree-based methods learn a hierarchical tree structure to partition the instances or labels into different groups so that similar ones can be in the same group.
The whole set, initially assigned to the root node, is partitioned into a fixed number k subsets corresponding to k child nodes of the root node. The partition process is repeated until a stopping condition is satisfied on the subsets. In the prediction phase, the input instance passes down the tree until it reaches the leaf node. For an instance tree, the prediction is given by the classifier
trained on the leaf instances. For a label tree, the prediction of a given label is the probability determined by the
traversed node classifiers from the root node to the leaf
node.

FastXML \cite{prabhu2014fastxml} learns a tree structure over the feature space by optimizing the normalized Discounted Cumulative Gain (nDCG). A binary classifier is trained for each internal node. The prediction for a given instance is the label distribution which is computed over the training instances in the corresponding leaf node.
Based on FastXML, PfastreXML \cite{jain2016extreme} introduces propensity scored losses which  prioritize predicting the
few relevant labels over the large number of irrelevant ones and promotes the prediction of infrequent, but rewarding tail labels.
Parabel \cite{prabhu2018parabel} learns a ensemble of three label trees. The label tree is trained by recursively partitioning the labels into two balanced
groups. The leaf nodes contain linear one-vs-all
classifiers, one for each label in the leaf. The one-vs-all classifiers are used to compute the 
probability of the corresponding labels relevant to the test
point.

CRAFTML \cite{siblini2018craftml} introduces a random forest-based method with a fast partitioning strategy. At first, it randomly projects both the feature and label vectors into low-dimensional vectors. Then a k-means based partitioning method splits the instances into k temporary subsets from the low-dimensional label vectors. A multi-class classifier in the internal node can be trained for its relevant subset from the low-dimensional feature vectors. 
Bonsai \cite{khandagale2019bonsai} develops a generalized label representation by combining the input and output representations. 
Then it constructs a shallow
tree architecture through the k-means clustering. Bonsai has demonstrated  fast speed to deal with an extremely large dataset.
ET \cite{wydmuch2018no} adopts the same architecture as FastText \cite{joulin2017bag}, but it employs the Probabilistic Label Tree as the output layer to deal with the multi-label classification instead of the Hierarchical Softmax. ET has the advantages of small model size and efficient prediction time. However, since the architecture of FastText cannot capture the rich contextual information of the text, the prediction accuracy cannot be high.

\textbf{Deep learning  approaches.} In contrast to BOW used by the traditional methods 
as the text representation, the raw text has been effectively utilized by deep learning models to learn the dense representation, which can capture the contextual and semantic information of the text.
Based on CNN-Kim \cite{kim2014convolutional}, XML-CNN \cite{liu2017deep} learns a number of feature representations by passing the text through convolution layers. A hidden bottleneck layer is added between the pooling and output layer to improve the prediction accuracy and reduce the model size. However, since the output layer of XML-CNN is a linear structure, it can be inefficient when applied to a case with millions of labels.

Inspired by the information retrieval (IR) perspective, X-BERT \cite{WeiCheng2019extreme} presents a method in three steps. At first, it builds the label indexing system by partitioning the label set into k clusters. Then it fine-tunes the Bidirectional Encoder Representations from
Transformers (BERT) \cite{devlin2019bert} model to match label clusters for the input texts. In the last step, the linear classifiers are trained to rank the labels in the corresponding cluster. X-BERT fine-tunes the BERT model successfully and has shown significant improvement in terms of prediction accuracy. However, X-BERT is not an entire end-to-end deep learning model, so it leaves potential room to improve the results.
AttentionXML \cite{you2019attentionxml} captures the sequential information of the text  by the Bidirectional Long Short-term Memory (BiLSTM) model and the Attention mechanism. A
shallow and wide probabilistic label tree is proposed to handle the large scale of labels. AttentionXML achieves outstanding performance on both accuracy and efficiency. However, an ensemble of three probabilistic label trees is adopted to improve the prediction accuracy, which is unusual in deep learning approaches.

\section{APLC-XLNet}
\label{section:aplc_xlnet}

\begin{figure}[t]
\centerline{\includegraphics[width=\columnwidth]{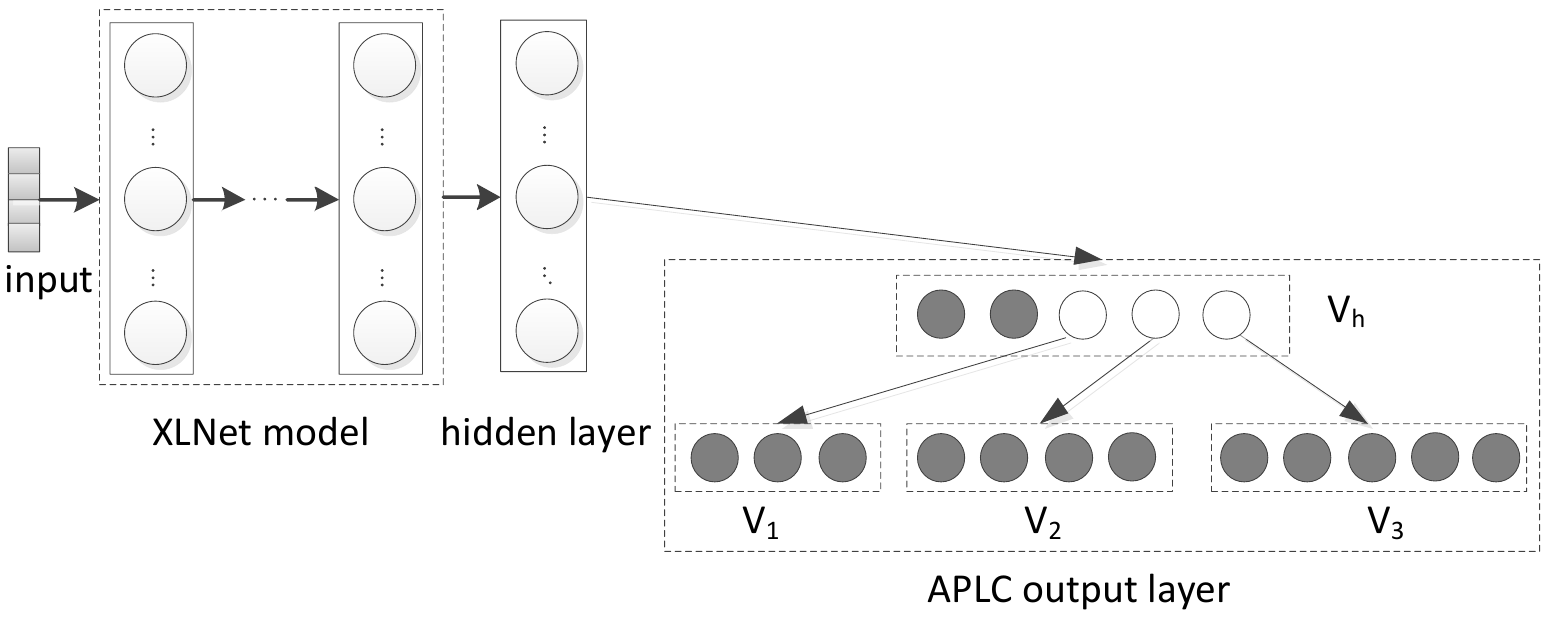}}
\caption{Architecture of the proposed APLC-XLNet model. $V$ denotes the label cluster in APLC.}
\label{fig:aplc_xlnet}
\vskip -0.1in
\end{figure}

In this section, we elaborate on our APLC-XLNet model for the XMTC problem. Our model architecture has three components: the XLNet module, the hidden layer and the APLC output layer (Figure \ref{fig:aplc_xlnet}).

\subsection{XLNet module}
Recently, there have been many significant advances in transfer learning for NLP, most of which are based on  
language modeling \cite{devlin2019bert,radford2018improving, conneau2019cross, yang2019xlnet}.  Language modeling is the task of predicting the next word in a sentence given previous 
words. XLNet \cite{yang2019xlnet} is a generalized autoregressive 
pretraining language model that captures bidirectional contexts by maximizing the expected likelihood over
all permutations of the factorization order. Its permutation language modeling objective can be expressed as follows:
\begin{equation}
 \max_{\theta} E_{z\sim Z_T}\left[ \sum_{t=1}^{T} log \: p_{\theta} \left( x_{z_t} | x_{z < t} \right)   \right]
\end{equation}
where $\theta$ denotes the parameters of the model, $Z_T$ is the set of all possible permutations of the sequence with length T, $z_t$ is the t-th element and $z < t $ deontes the preceding $ t-1$ elements in a permutation $z$. Essentially, for a text sequence $x$, a factorization order $z$ can be sampled  at one time, then the likelihood $p_\theta ( x  ) $ can be obtained by  decomposing this order. 

In addition to  permutation language modeling, XLNet adopts  the Transformer XL \cite{dai2019transformer} as its base architecture. XLNet pretrained the language model on a large corpus of text data, and then fine-tuned the pretrained model on downstream tasks. While the pretrained model has achieved state-of-the-art performance on many downstream tasks including sentence classification, question answering and
sequence tagging, it remains  unexplored for the application of extreme multi-label text classification.

We adopt the pretrained XLNet-Base model in our model architecture and fine-tune it to learn the powerful representation of the text. Following the approach of XLNet \cite{yang2019xlnet}, we begin with the pretrained XLNet model with one embedding layer, then the module consisting of 12 Transformer blocks.  We take the final hidden vector of the last token corresponding to the special [CLS] token as the text representation.

\subsection{Hidden layer}
APLC-XLNet has one fully connected layer between the pooling layer of XLNet and APLC output layer. We choose to set the number of neurons in this layer as a hyperparameter $d_{h}$. When the number of output labels is not large, $d_h$ can be the same as the pooling layer to obtain the best text representation. On the other hand, when handling  the case of extreme labels, it can be less than the pooling layer to largely reduce the model size and make the computation more efficient. In this case, this layer can be referred to as a bottleneck layer, as the number of neurons is less than both the pooling layer and output layer.

\subsection{Adaptive Probabilistic Label Clusters}
\begin{figure}[t]
\vskip -0.1in
\begin{center}
\centerline{\includegraphics[width=\columnwidth]{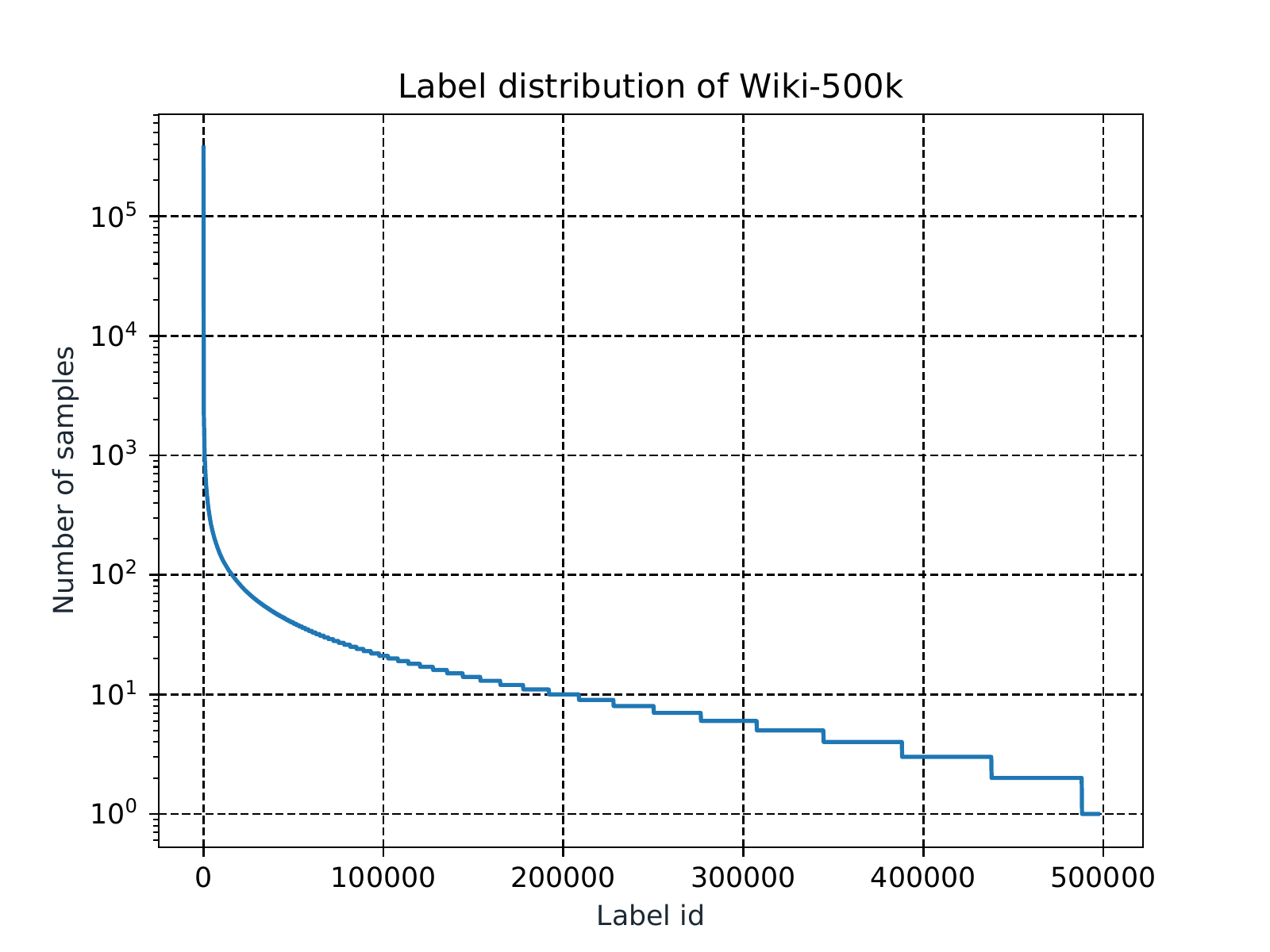}}

\vskip -0.1in

\caption{Label distribution of dataset Wiki-500k follows Zipf's Law. Label ids are sorted in descending order by the number of occurrence in the dataset. }
\label{fig:wiki500k}
\end{center}
\vskip -0.4in
\end{figure}

\textbf{Motivation.} 
In extreme classification, the distribution of the labels notoriously follows Zipf's Law. Most of the probability mass is covered by only a small fraction of the label set. In one benchmark dataset, Wiki-500k, the frequent labels account for 20\% of the label vocabulary, but they cover about 75\% of probability mass (as shown in Figure \ref{fig:wiki500k}).
Similar to Hierarchical Softmax \cite{mikolov2011extensions} and Adaptive Softmax \cite{grave2017efficient}, this attribute can be exploited to reduce the computation time. 

\begin{figure}[t]
\vskip +0.1in
\begin{center}
\centerline{\includegraphics[width=\columnwidth]{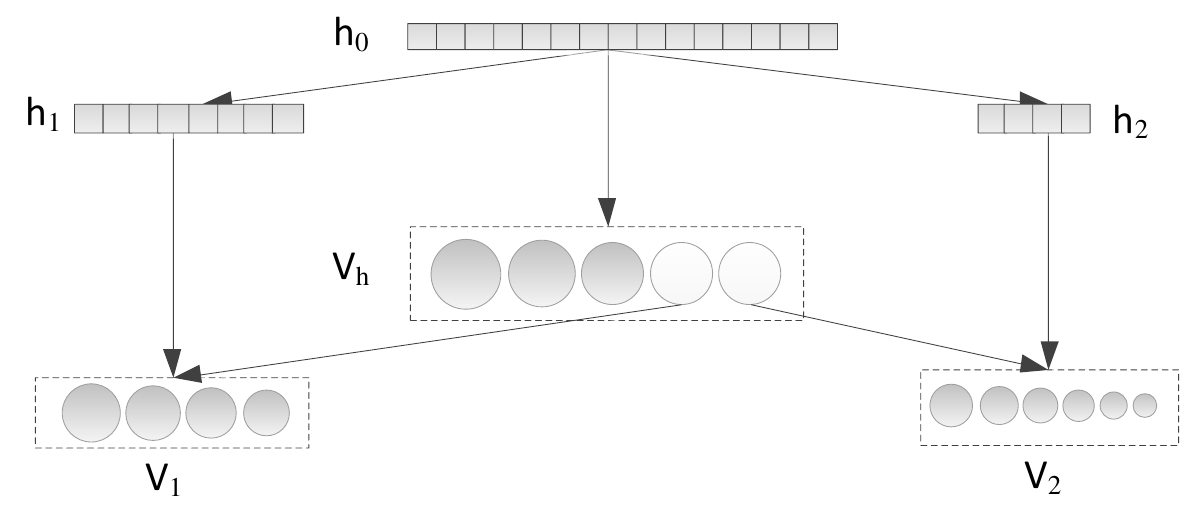}}
\vskip -0.0in

\caption{Architecture of the 3-cluster APLC. $h$ denotes the hidden state. $V_h$ denotes the head cluster. $V_1$ and $V_2$ denote the two tail clusters. The size of the leaf node indicates the frequency of the label.}

\label{fig:aplc}
\end{center}
\vskip -0.5in
\end{figure}

\textbf{Architecture.}
We partition the label set into a head cluster and several tail clusters, where the head cluster consists of the most frequent labels and the infrequent labels are grouped into the tail clusters (Figure \ref{fig:aplc}). The head cluster containing a small fraction of labels corresponds to less computation time, so it would improve the efficiency of computation greatly by accessing it frequently. The structure of clusters can be developed in two different ways. One way is to generate a 2-level tree \cite{mikolov2011extensions}, while the other is to keep the head cluster as a short-list in the root node \cite{le2011structured, grave2017efficient}. Empirically, it leads to a considerate decrease of performance to put all the clusters in the leaves. Like the Adaptive Softmax  \cite{grave2017efficient}, we choose to put the head cluster in the root node. To further reduce the computation time, we adopt decreasing dimensions of hidden state as the inputs to the clusters. For the head cluster accessed by the classifier frequently, a large dimension of hidden state can keep the high prediction accuracy. For the tail clusters, since the classifier would access them infrequently, we decrease the dimension through dividing by the factor $q$ ($q \geq 1 $). In this way, the model size of the clusters can be significantly reduced, yet the model can still maintain good performance.

\textbf{Objective function.}  We assume the label set is partitioned as $V = V_h \cup V_1\cup \ldots \cup V_K$, where  $V_h$  is the head cluster, $V_i$  is the i-th tail cluster, \: $ V_h \cap  V_i = \emptyset  \: \;  \text{and}  \: \; V_i \cap  V_j = \emptyset, \: \text{if}\: i \neq j$, $K$ is the number of tail clusters. By the chain rule of probability, the probability of one label can be expressed as follows:
\begin{equation}
\label{eqn:equation_2}
p(y_j|x)=\begin{cases}
  p(y_j|x), & \text{if $y_j \in V_h$}\\
  p(V_t|x) p(y_j|V_t,x), & \text{if $y_j \in V_t$}
  \end{cases}
\end{equation}
where $x$ is the feature of one sample, $y_j$ is the j-th label, $V_t$ is the t-th tail cluster.

During training, we access the head cluster,  where we compute the probability of each label for each training sample. In contrast, we access the tail cluster $V_t$, where we compute the probability of each label only if there is a positive label of the training sample located in $V_t$.
Let $Y_i$ be the set of positive labels of the i-th instance, $V_{y_k}$ is the corresponding cluster of the k-th label in $Y_i$, the set of clusters $S_i $ corresponding to $Y_i$ can be expressed as follows: 
\begin{equation}
S_i = \bigcup_{ y_{k} \in Y_i}V_{y_k}
\end{equation}
Note that $S_i $ can either contain or not contain the head cluster
$V_h $, but we need to access $V_h $ for each training sample. We add $V_h $ into $S_i $ and the set of clusters $\hat {S_i}$ corresponding to $Y_i $ 
can be expressed as follows:
\begin{equation}
\hat {S_i} = S_i \bigcup V_h = (\bigcup_{ y_{k} \in Y_i}V_{y_k}) \bigcup V_h
\end{equation}
Let $\hat {Y_i}$ be the label set corresponding to $\hat {S_i}$. We use $L_i$ to denote the cardinality of $\hat {Y_i}$, and $I_i$ denote the set of label indexes of $\hat {Y_i}$. The objective loss function of APLC for multi-label classification can be expressed as follows:
%
\begin{multline}
J(\theta)=- \frac{1}{\sum_{i=1}^{N}L_i} \sum_{i=1}^{N} \sum_{  j \in I_i}( y_{ij}log  \: p ( y_{ij}  )  + \\ 
 ( 1-y_{ij} ) log  \: (1 - p (y_{ij} ) ) )
\end{multline}
%
where N is the number of samples, $p(y_{ij})$ is the predicted probability computed from Equation \ref{eqn:equation_2}, $y_{ij} \in \{0,1\}$ is the true value, index i and j denote the i-th sample and j-th label respectively.

\textbf{Model size.} Let us conduct the analysis on the model size of APLC.
Let $d$ denote the dimension of the hidden state of $V_h$, $q\ (q \geq 1) $ denotes the decay factor, $l_h$ and $ l_i$ denote the cardinality of the head cluster and i-th tail cluster.
The number of parameters $N_{par}$ of APLC can be expressed as follows:
\begin{equation}
\label{eqn:equation_6}
N_{par} = d(l_h + K) + \sum_{i=1}^{K}  \frac{d}{q^i} (d +  l_i) 
\end{equation}
In practice, $K \ll l_h$ and $d \ll l_i$, so Equation \ref{eqn:equation_6} can be expressed as:
\begin{equation}
\label{eqn:equation_7}
N_{par} \approx d l_h + \sum_{i=1}^{K}  \frac{d}{q^i} l_i = d \sum_{i=0}^{K}  \frac{l_i}{q^i} 
\end{equation}
where $l_0$ denotes ${l_h}$. 
As shown in Equation \ref{eqn:equation_7}, the last tail cluster has the smallest coefficient.  Let us consider the case when $d$ and $q$ are fixed; the strategy to reduce the model size is to assign the large fraction of labels to the tail clusters.

Let us make a comparison with the original linear output layer. The model size of the linear structure can be expressed as follows:
\begin{equation}
\label{eqn:equation_8}
N_0 = d L
\end{equation}
where $L$ denotes the cardinality of the label set.  Combining  Equation \ref{eqn:equation_7} and Equation \ref{eqn:equation_8}, we have the expression of the ratio between them:  
\begin{equation}
\frac{N_0}{N_{par}} = \frac {d L}{d \sum_{i=0}^{K}  \frac{l_i}{q^i}} =  \frac {L}{ \sum_{i=0}^{K}  \frac{l_i}{q^i}} , \: \text{where} \: L = \sum_{i=0}^{K} l_i  
\end{equation}

\textbf{Computational complexity.}
The expected computational cost $ C$ can be described as follow:
\begin{equation}
\label{eqn:equation_10}
C = C_h + \sum_{i=1}^{K} C_i 
\end{equation}
where $C_h$ and $C_i$ denote the computational cost of the head cluster and the i-th tail cluster, respectively. We let $N_b$ to denote the batch size, and $p_i $ to denote the probability that at least one of the positive labels of a batch of samples is in the tail cluster $V_i$ and the model would access $V_i$. We have the following expression:
\begin{equation}
\begin{split}
C &= O(N_b d (l_h+K)) + O (\sum_{i=1}^{K}p_i N_b \frac{d}{q^i}(l_i + d)) \\
& = O(N_b d (l_h+K+\sum_{i=1}^{K}p_i \frac{l_i+d}{q^i}))  
\end{split}
\end{equation}
In practice, $K \ll l_h $ and $d \ll l_i$, so the computational cost $C$ can be expressed as follows:
\begin{equation}
\label{eqn:equation_12}
C = O(N_b d (l_h+\sum_{i=1}^{K}p_i \frac{l_i}{q^i}))  
\end{equation}
Let us consider the case that we have partitioned the label set into clusters where the cardinality of each cluster is fixed. In Equation \ref{eqn:equation_12}, all the values are fixed except the probability $p_i$ of each tail cluster. In order to have $p_i$ take a small value, we should assign the most frequent labels into the head cluster. On the other hand, since we have assigned decreasing dimensions of hidden state to the tail clusters, we should partition the labels by decreasing frequency to the tail clusters to obtain high prediction accuracy.

\begin{table*}[t]
\begin{center}
\vskip -0.1 in
\caption{Statistics of datasets. $N_{train}$ is the number of training samples, $N_{test}$ is the number of test samples, $D$ is the dimension of the feature vector, $L$ is the cardinality of the label set, $\bar{L}$ is the average number of labels per sample, $\hat{L}$ is
the average samples per label, $\overline W_{train}$ is the average number of words per training sample and $\overline W_{test}$ is the average
number of words per test sample. }
\vskip +0.1in
\label{table:dataset}
\begin{tabular}{c r r r r r r r r}
\hline
Dataset & $N_{train}$ & $N_{test}$ & $D$ & $L$ & $\bar{L}$ & $\hat{L}$ &
$\overline W_{train}$ & $\overline W_{test}$\\
\hline
EURLex-4k & 15,539 & 3,809 & 186,104 & \textbf{3,956} & 5.30 & 20.79 & 1,248.58 & 1,230.40  \\
AmazonCat-13k & 1,186,239 & 306,782 & 203,882 & \textbf{13,330} & 5.04 & 448.57 & 246.61 & 245.98  \\
Wiki10-31k & 14,146 & 6,616 & 101,938 & \textbf{30,938} & 18.64 & 8.52 & 2,484.30 & 2,425.45  \\
Wiki-500k & 1,646,302 & 711,542 & 2,381,304 & \textbf{501,069} & 4.87 & 16.33 & 750.64 & 751.42  \\
Amazon-670k & 490,449 & 153,025 & 135,909 & \textbf{670,091} & 5.45 & 3.99 & 247.33 & 241.22  \\
\hline
\end{tabular}
\vskip -0.1in
\end{center}
\end{table*}
Let us also make a comparison with the original linear output layer. The expected computation cost of the linear structure can be expressed as follow:
\begin{equation}
\label{eqn:equation_13}
C_0 = O(N_b d L) 
\end{equation}
Combining Equation \ref{eqn:equation_12} and Equation \ref{eqn:equation_13}, the ratio between them can be expressed as follows:

\begin{equation}
\begin{split}
\frac{C_0}{C} &= \frac {O(N_b d L)}{O(N_b d (l_h+\sum_{i=1}^{K}p_i \frac{l_i}{q^i}))} \\
& = \frac {L}{l_h+\sum_{i=1}^{K}p_i \frac{l_i}{q^i}},  \: \text{where} \: L = l_h + \sum_{i=1}^{K} l_i 
\end{split}
\end{equation}
\begin{table}
\begin{center}
\vskip -0.25in
\caption{Implementation details of APLC.  $d_{h}$ is the dimension of the input hidden state, $q$ is the factor by which the dimension of hidden state for the tail cluster  decreases, $N_{cls}$ is the number of clusters and $P_{num}$ is the proportion for which the number of  labels in each cluster accounts. }
\vskip +0.1in
\label{table:aplc}

\begin{adjustbox}{width=0.48\textwidth}
\begin{tabular}{c c c c c }
\toprule
Dataset & $d_{h}$ & $ q$ & $N_{cls}$ & $P_{num} $  \\
\midrule
EURLex-4k & 768 & 2 & 2& 0.5, 0.5 \\
AmazonCat-13k & 768 & 2 & 2 & 0.5, 0.5 \\
Wiki10-31k & 768 & 2 & 2 & 0.5, 0.5 \\
Wiki-500k & 768 & 2 & 3 & 0.33, 0.33, 0.34  \\
Amazon-670k & 512 & 2 & 4 & 0.25, 0.25, 0.25, 0.25\\
\bottomrule
\end{tabular}

\end{adjustbox}
\end{center}
\vskip -0.45in
\end{table}
\raggedbottom
\section{Techniques to train
APLC-XLNet}
\label{section:techniques}
\subsection{Discriminative fine-tuning}
APLC-XLNet consists of three modules, the XLNet module, the hidden layer and the APLC output layer. When dealing with the extreme classification problem with millions of labels, the number of parameters in the APLC output layer can be even greater than the XLNet model. Hence, it is a significant challenge to train such a large model. We adopt the discriminative fine-tuning method \cite{howard2018universal} to train the model. Since the pretrained XLNet model has captured the universal information for the downstream tasks, we should assign the learning rate $\eta _ x $ a small value.  We set a greater value to $ \eta _ a $ for the APLC output layer to motivate the model to learn quickly.
Regrading to  $ \eta _ h $ for the intermediate hidden layer, we assign it a value  between the ones of XLNet model and APLC layer. 
Actually, in our experiments, we found this method was necessary to train the model effectively. It is infeasible to train the model with the same learning rate for the entire model.

\subsection{Slanted triangular learning rates}
Slanted triangular learning rates \cite{howard2018universal} is an approach of using a dynamic learning rate
to train the model. The objective is to motivate the model to converge quickly to the suitable space at the beginning and then refine the parameters.  Learning rates are first increased linearly, and then decayed gradually according to the strategy. The learning rate $\eta $ can be expressed as follows:\\
\begin{equation}
\eta  =\begin{cases}
  \eta_0 \frac{t}{t_w}, & \text{if $t \leq t_w$}\\
 \eta_0 \frac{t_a - t}{t_a - t_w}, & \text{if $t > t_w$}
  \end{cases}
\end{equation}
where $\eta_0 $ is the original learning rate, t denotes the current training step, the hyperparameter $t_w$ is the warm-up step threshold, and $t_a$ is the total number of training steps. 

\begin{table}[t]

\vskip -0.25in
\begin{center}
\caption{Hyperparameters for training the model. $L_{seq}$ is the length of input sequence. $ \eta_x$, $\eta_h$, and $\eta_a$ denote the learning rate of the XLNet model, the hidden layer and APLC layer, respectively. $N_b$ is the batch size and $N_e$ is the number of training epochs. }
\vskip +0.1in
\label{table:hyperparameter}
\begin{adjustbox}{width=0.48\textwidth}
\begin{tabular}{c c c  c c c c }
\toprule
Dataset & $L_{seq}$ & $  \eta _ x $ & $ \eta _ h$ & $ \eta _ a$ & $N_b$& $N_e$\\
\midrule
EURLex-4k & 512 & 5e-5 & 1e-4 & 2e-3 & 12 & 8  \\
AmazonCat-13k & 192 & 5e-5 & 1e-4 & 2e-3 & 48 & 8  \\
Wiki10-31k & 512 & 1e-5 & 1e-4 & 1e-3 & 12 & 6  \\
Wiki-500k & 256 & 5e-5 & 1e-4 & 2e-3 & 64 & 12  \\
Amazon-670k & 128 & 5e-5 & 1e-4 & 2e-3 & 32 & 25  \\
\bottomrule
\end{tabular}
\end{adjustbox}
\vskip -0.08in
\end{center}
\vskip -0.2in
\end{table}

\section{Experiments}
\label{section:experiments}
In this section, we report the performance of our proposed method on standard datasets and compare it against  state-of-the-art baseline approaches.

\textbf{Datasets.} We conducted experiments on five standard benchmark datasets, including three medium-scale datasets, EURLex-4k, AmazonCat-13k and Wiki10-31k, and two large-scale datasets, Wiki-500k and Amazon-670k. Table \ref{table:dataset} shows the statistics of these datasets. The term frequency–inverse document frequency (tf–idf) features for the five datasets are publicly available
at the Extreme classification Respository\footnote{\url{http://manikvarma.org/downloads/XC/XMLRepository.html}}. We used the raw text of 3 datasets, including AmazonCat-13k, Wiki10-31k and Amazon-670k, from the the Extreme classification Respository. We obtained the raw text of EURLex\footnote{\url{http://www.ke.tu-darmstadt.de/resources/eurlex/eurlex.html}} and Wiki-500k\footnote{\url{https://drive.google.com/drive/folders/1KQMBZgACUm-ZZcSrQpDPlB6CFKvf9Gfb}} from the public websites.

\begin{table*}[t]
\centering

\vskip -0.1in
\caption{Comparisons between APLC-XLNet and state-of-the-art baselines. The results of the baselines are cited from the original papers except the baseline AttentionXML. The best result among all the methods is in bold.  }
\label{table:results}
\vskip +0.1in
\begin{adjustbox}{width=1\textwidth}
\begin{tabular}{c c c c c c c c c c c}
\toprule
Dataset &  & SLEEC &  AnnexML & DisMEC & PfastreXML& Parabel & Bonsai & XML-CNN  &
 AttentionXML   & APLC-XLNet\\
\hline
 & P@1 & 79.26 & 79.66 & 82.40 & 75.45 & 81.73 & 83.00 & 76.38 & 87.14&  \textbf{87.72} \\
EURLex-4k & P@3 & 64.30 & 64.94 & 68.50 & 62.70 & 68.78 & 69.70 & 62.81 & \textbf{75.18}&  74.56 \\
 & P@5 & 52.33 & 53.52 & 57.70 & 52.51 & 57.44 & 58.40 & 51.41 & \textbf{62.58} &  62.28 \\
 \hline
 & P@1 & 90.53 & 93.55 & 93.40 & 91.75 & 93.03 & 92.98 & 93.26 & 92.62&  \textbf{94.56} \\
AmazonCat-13k & P@3 & 76.33 & 78.38 & 79.10 & 77.97 & 79.16 & 79.13 & 77.06 & 77.56 &  \textbf{79.82} \\
 & P@5 & 61.52 & 63.32 & 64.10 & 63.68 & 64.52 & 64.46 & 61.40 & 62.74 &  \textbf{64.60} \\
 
 \hline
 & P@1 & 85.88 & 86.50 & 85.20 & 83.57 & 84.31 & 84.70 & 84.06 & 86.04&  \textbf{89.44} \\ 
Wiki10-31k & P@3 & 72.98 & 74.28 & 74.60 & 68.61 & 72.57 & 73.60 & 73.96 & 77.54&  \textbf{78.93} \\
 & P@5 & 62.70 & 64.19 & 65.90 & 59.10 & 63.39 & 64.70 & 64.11 & 68.48 &  \textbf{69.73} \\ 
  \hline
 & P@1 & 53.60 & 63.86 & 70.20 & 56.25 & 68.52 & 69.20 & 59.85 & 72.62& \textbf{72.83} \\ 
Wiki-500k & P@3 & 34.51 & 40.66 & 50.60 & 37.32 &  49.42 & 49.80 & 39.28 & \textbf{51.02}& 50.50 \\
 & P@5 & 25.85 & 29.79 & \textbf{39.70} & 28.16 &  38.55 & 38.80 & 29.81 & 39.41& 38.55 \\
 
   \hline
 & P@1 & 35.05 & 42.08 & 44.70 & 39.46 & 44.89 & \textbf{45.50} & 35.39 & 45.45& 43.46 \\ 
Amazon-670k & P@3 & 31.25 & 36.65 & 39.70 & 35.81 & 39.80 & 40.30 & 31.93 & \textbf{40.63}& 38.82 \\
 & P@5 & 28.56 & 32.76 & 36.10 & 33.05 & 36.00 & 36.50 & 29.32 & \textbf{36.92}& 35.32 \\
 
\bottomrule
\end{tabular}
\end{adjustbox}
\end{table*}

\textbf{Implementation details.} 
We need to use a specific tokenizer to preprocess the raw text,  which is based on  SentencePiece tokenizer \cite{kudo2018sentencepiece}. During tokenization, each word in the sentence was broken apart into small tokens. Then we  chose sentence length $L_{seq} $,  padded and truncated every input sequence to be the same length.
Table \ref{table:aplc} shows the implementation details of APLC. For medium-scale datasets, we choose to evenly partition the label set into two clusters. For the large-scale datasets Wiki-500k and Amazon-670k, we evenly divide the label set into three and four clusters respectively. To further reduce the model size for the large-scale dataset, Amazon670k, we set the dimension of hidden state $d_{h}$ to be 512. The decay factor $q $ is 2 for all five datasets.

Table \ref{table:hyperparameter} shows the hyperparameters for training the model. There are several factors to consider for setting the sequence length $L_{seq} $. First, a long sequence contains more contextual information, which is beneficial for the model to learn a better text representation. Second,
it is linearly proportional to the computational time. For datasets with a small number of training samples, we set $L_{seq} $ to be the maximum value 512. For datasets with a large number of training samples, we choose smaller values for $L_{seq} $. We choose the AdamW optimizer and set different learning rates. The learning rate $ \eta_x $ for the XLNet model is at the magnitude of 1e-5, $ \eta_a $ for the APLC layer is at the magnitude of 1e-3, and  $ \eta_h $
for the intermediate hidden layer is between $ \eta_x $ and  $ \eta_a $. The warm-up step $ t_w$ is 0 for all five datasets.

\textbf{Evaluation Metrics.} 
We choose the widely used P@k as the evaluation metric, which represents the prediction accuracy by computing the precision of top k labels.  $P@k$ can be defined as follows:
\begin{equation}
P@k = \frac{1}{k}\sum_{i\in rank_k (\hat y)}y_i
\end{equation}
where $ \hat y$ is the prediction vector, $i$ denotes the index of the i-th highest element in $ \hat y$ and $y \in \{0,1\}^L$.

\textbf{Baselines.}
Our method is compared to  state-of-the-art baselines, including  one-vs-all in DisMEC \cite{babbar2017dismec}, three tree-based approaches, PfastreXML \cite{jain2016extreme}, Parabel \cite{prabhu2018parabel}  and Bonsai \cite{khandagale2019bonsai}, two embedding-based approaches, SLEEC \cite{bhatia2015sparse} and AnnexML \cite{tagami2017annexml}, and two deep learning approaches, XML-CNN \cite{liu2017deep} and AttentionXML \cite{you2019attentionxml}. We ran the source code of AttentionXML on the five datasets used in this paper, which are different from the datasets in AttentionXML. Note that AttentionXML leverages an ensemble of three trees to improve the performance. For the sake of a fair comparison, we only choose the results produced by one tree. We have released the preprocessed datasets for AttentionXML publicly on GitHub to reproduce the results of AttentionXML in this paper.  

\textbf{Performance comparison.} Table \ref{table:results} shows the experimental results of APLC-XLNet and the state-of-the-art baselines over the five datasets. Following the previous work on XMTC, we consider the top k prediction precision, $ P@1$, $ P@3$ and $ P@5$. At first, we compare APLC-XLNet with two embedding-based approaches, SLEEC and AnnexML. AnnexML performs better than SLEEC over all five datasets. However, APLC-XLNet outperforms AnnexML over all five datasets. The improvements are significant on the dataset EURLex-4k, with an increase of about 8 percent, 10 percent and 10 percent on $ P@1$, $ P@3$ and $ P@5$ respectively.
The 1-vs-all approach, DisMEC has the best performance among all the methods on dataset Wiki-500k in terms of $ P@5$. APLC-XLNet outperforms DisMEC on three datasets, while its performance is slightly worse than DisMEC on dataset Amazon-670k, with a drop of 1 percent.
Bonsai has the best performance on four datasets among the three tree-based approaches.
The performance of Parabel outperforms Bonsai on the dataset AmazonCat-13k. Our approach outperforms the three tree-based approaches by a large margin on three datasets, EURLex-4k, AmazonCat-13k and Wiki10-31k. The deep learning approach AttentionXML has the best performance on dataset Amazon-670k among all methods.
APLC-XLNet outperforms AttentionXML on two datasets AmazonCat-13k and Wiki10-31k. Note that the three deep learning approaches take the raw text as the input, and can make use of the contextual and semantic information of the text. However, they utilize different models to learn the text representation.

\begin{figure}[t]
\vskip -0.1in
\begin{center}
\centerline{\includegraphics[width=\columnwidth]{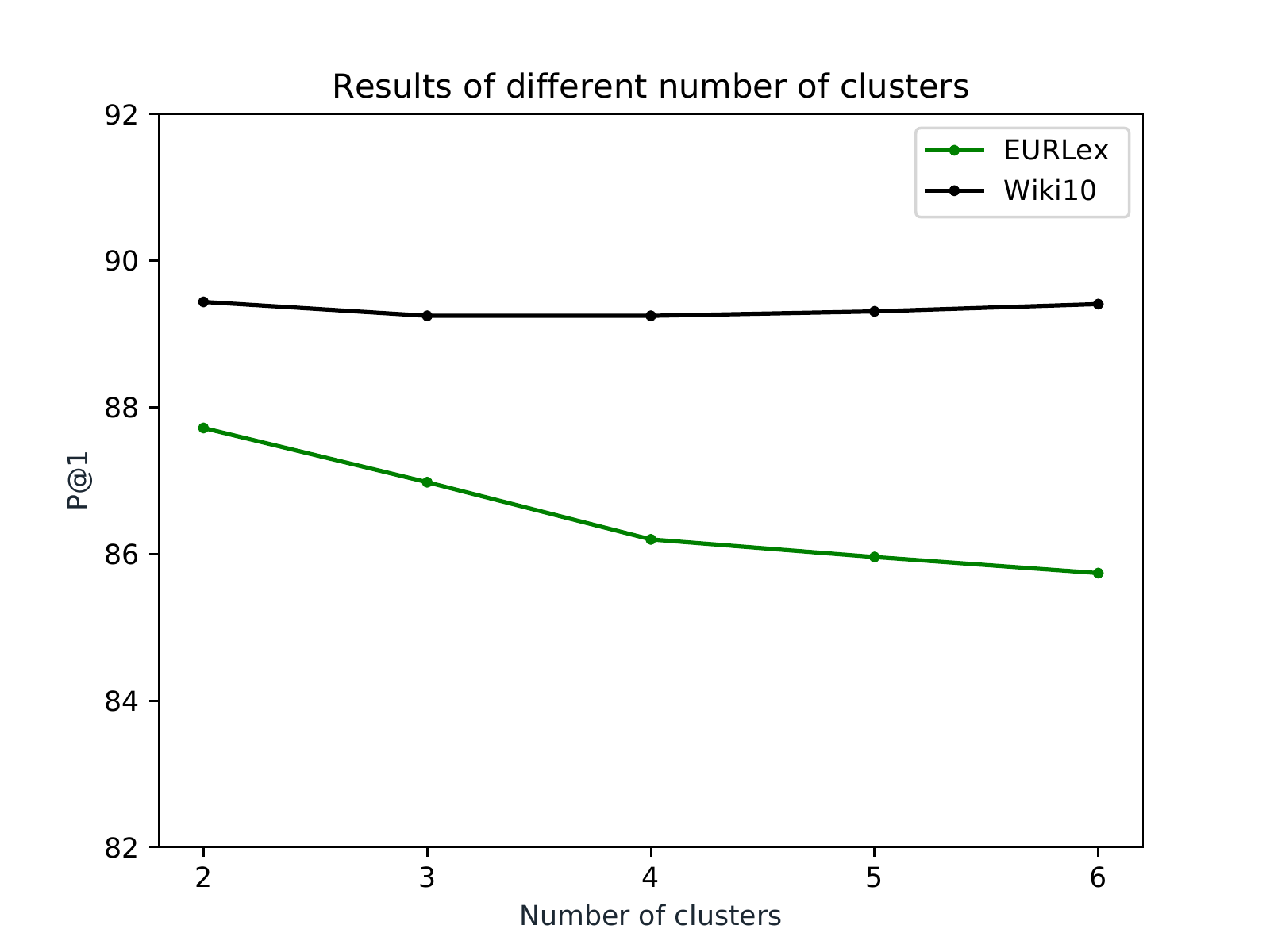}}
\caption{Precision $P@1$ on dataset EURLex and Wiki10, as a function of the number of clusters.}
\label{fig:cluster}
\end{center}
\vskip -0.25in
\end{figure}

\textbf{Ablation study.}
We perform an ablation study to understand the impact of different design choices of APLC  based on two datasets, EURLex and Wiki10, with diverse characteristics. Specifically, there are two factors we consider: (1) the impact of the number of clusters, and (2) the impact of the method to partition the label set.

To answer the first question, we assume  when the number of clusters is given, the label set is evenly partitioned into each cluster. The other parameters are the same setting as Table \ref{table:aplc} and Table \ref{table:hyperparameter}. We plot Figure \ref{fig:cluster}, precision $P@1$ as a function of the number of clusters $N_{cls}$. Dataset EURLex has the highest $P@1$ 87.72  when $N_{cls}$ is 2. As the value of $N_{cls}$ increases, the precision $P@1$ decreases gradually. When $N_{cls}$ reaches up to 6, the precision $P@1$ is 85.72, with a 2 percent drop. For dataset Wiki10, as the value of $N_{cls}$ increases, the precision $P@1$ decreases slightly. We argue that the large number of clusters tends to harm the performance of the model; however, the degree of impact depends on the characteristics of the dataset.

\begin{figure}[t]
\vskip -0.1in
\begin{center}
\centerline{\includegraphics[width=\columnwidth]{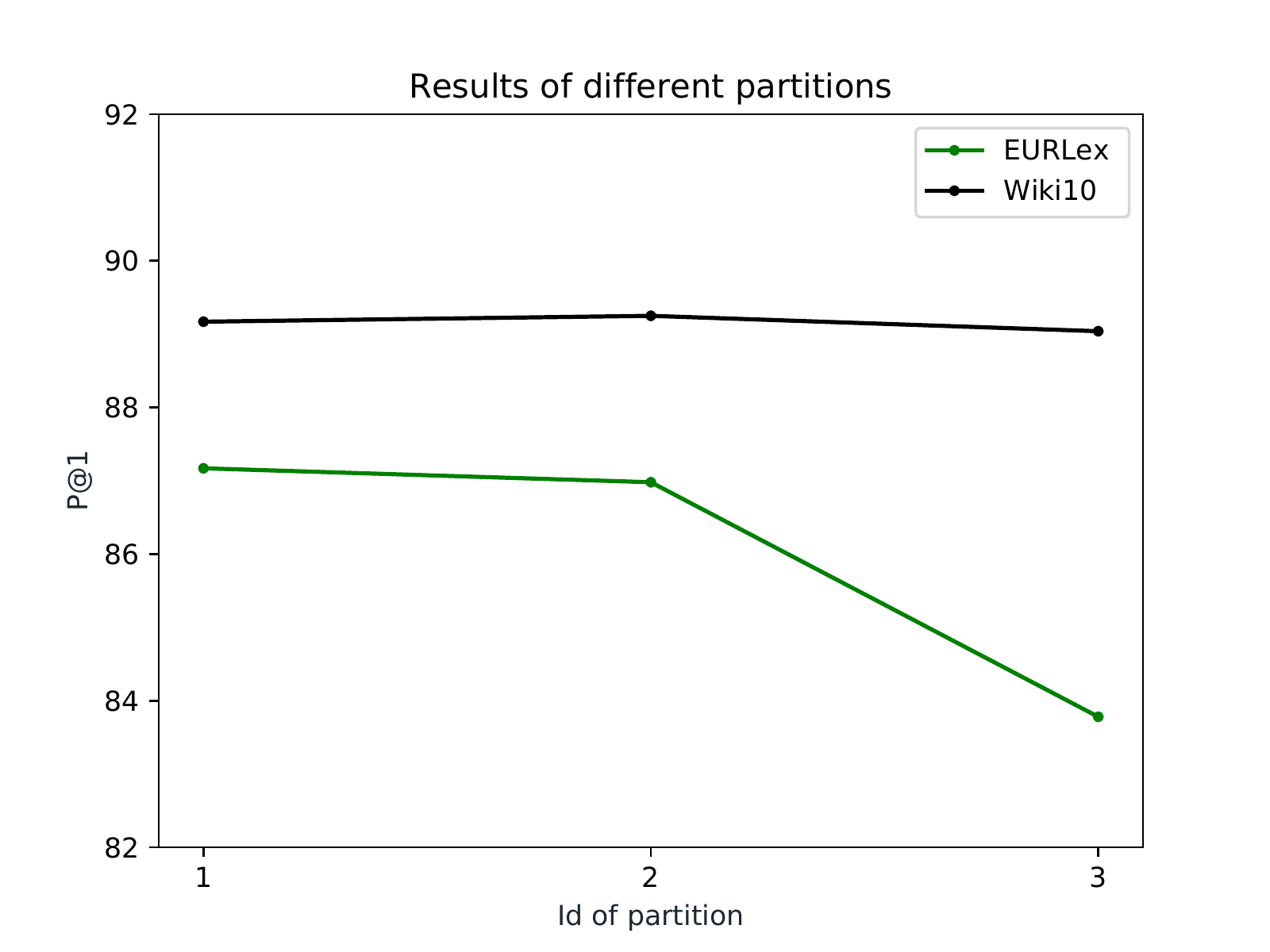}}
\caption{Precision $P@1$ of different partitions on dataset EURLex and Wiki10. Id 1, 2 and 3 denote the partition (0.7, 0.2, 0.1), (0.33, 0.33, 0.34) and (0.1, 0.2, 0.7) respectively.}
\label{fig:partition}
\end{center}
\vskip -0.25in
\end{figure}

To answer the second question, we assume the number of clusters $N_{cls}$ is a fixed value, 3. Let $V_h$, $V_1$ and $V_2$ denote the head cluster, the first tail cluster and the second tail cluster, and $P_h$, $P_1$, $P_2$ denote the proportion for which the number of labels in the corresponding cluster accounts.  We have three different ways to partition the label set, corresponding three combinations, (0.7,0.2,0.1), (0.33,0.33,0.34) and (0.1,0.2,0.7) for ($P_h$,  $P_1$, $P_2$).  The other parameters have the same settings as Tables \ref{table:aplc} and \ref{table:hyperparameter}.  We plot Figure \ref{fig:partition}, the prediction precision $P@1$ of different partitions on dataset EURLex and  Wiki10. We observe that when more labels are partitioned into the head cluster, the precision  $P@1$ is higher on both datasets. This trend is more significant on dataset EURLex, as there is an about 3 percent difference between the first partition and the third partition. However, the impact on dataset Wiki10 is relatively small.

\section{Conclusion}
\label{conclusion}
In this paper, we have proposed a novel deep learning approach for the XMTC problem. In terms of prediction accuracy, the performance of our method has achieved new state-of-the-art results over four  benchmark datasets, which has shown that the text representation fine-tuned from the pretrained XLNet model is more powerful. Furthermore, we have proposed APLC to deal with extreme labels efficiently.  We  have carried out theoretical analysis on the model size and computation complexity for APLC.  The application of APLC is not limited to XMTC. We believe that APLC may be general enough to be applied to the extreme classification problem as the output layer, especially in tasks where the distributions of classes are unbalanced.

\newpage

\section*{Acknowledgements}

We would like to thank the anonymous reviewers for their insightful comments and  suggestions, and Shihao Ji for
feedback on the manuscript.


\bibliography{main}
\bibliographystyle{icml2020}





\end{document}